\documentclass[sigconf,nonacm]{acmart}
\usepackage{enumitem}
\usepackage{hyperref}
\begin{document}

\title[From Acceptance and Avoidance to Affective Viscosity: Designing Social Agents Intermediating in Human Relationships]{From Acceptance and Avoidance to Affective Viscosity: \\Designing Social Agents Intermediating in Human Relationships}

\author{Tomoko Yonezawa}
\orcid{0000-0001-8226-0815}
\affiliation{%
  \institution{Kansai University}
  \city{Takatsuki}
  \state{Osaka}
  \country{Japan}
}
\email{yone@kansai-u.ac.jp}

\renewcommand{\shortauthors}{Tomoko Yonezawa}

\begin{abstract}
As social agents become embedded in human relationships, they are expected not only to provide smooth assistance but also to handle rejection, discomfort, jealousy, persistence, and other forms of social tension. This paper connects two research directions: acceptance and avoidance in response to an agent's gaze, approach, and contact, and affective viscosity as a model of how negative affect accumulates, persists, is suppressed, and is expressed over time. We argue that avoidance should not be treated merely as a failure of interaction, but as a relational signal that indicates boundaries between humans and agents. Building on this view, affective viscosity provides a design perspective for agents that maintain, suppress, or express negative states depending on social relationships and interaction histories. We discuss several ongoing design cases, including triadic jealousy, discomfort suppression, patience timing, and spatiotemporal stickiness, and consider how these cases lead to broader questions of blame and accountable intervention. Rather than designing agents as simple scapegoats or artificial villains, we ask 
how agents can intermediate in problematic social situations
while making their intervention traceable, limited, contestable, and repairable.
\end{abstract}



\maketitle


\section{Introduction}
Social agents are often designed to be polite, helpful, and smoothly cooperative. However, agents embedded in human relationships may need to do more than maintain pleasant interaction. They may be approached or avoided, accepted or rejected, welcomed or perceived as intrusive. They may also need to express hesitation, discomfort, jealousy, or limits of patience when social tension arises. These reactions should not be treated only as breakdowns of interaction. They may indicate relational boundaries and provide important cues for how agents should intervene in social environments.

Our previous and ongoing work on acceptance and avoidance focuses on how users respond to an agent's gaze, approach, and contact, and how such responses can be estimated and reflected in the agent's behavior. This perspective treats avoidance not simply as a negative outcome, but as a signal that the agent has entered a boundary-sensitive region of interaction. The next step is to model how such boundary-sensitive states persist and transform over time. In particular, discomfort, jealousy, annoyance, and reluctance may accumulate, remain latent, leak through behavior, or suddenly become explicit. We refer to this temporal and relational property as affective viscosity.

This perspective also leads to questions of blame and responsibility. Once an agent detects social tension and intervenes, it may change the trajectory of the relationship. If the intervention fails, becomes annoying, or exposes a tension that participants preferred to ignore, the agent may become a target of blame. However, assigning blame to an agent should not simply mean placing an artificial scapegoat or villain in the environment. A more productive question is how to design agents whose interventions are traceable, limited, contestable, and repairable. In this sense, affective viscosity is not only a model of negative emotion, but also a basis for discussing how agents can take part in socially risky intervention.

Prior work on trust in automation and human--robot interaction has
emphasized that trust is not simply a positive attitude, but a basis
for appropriate reliance under uncertainty~\cite{lee2004trust,
hancock2011trust}. Reviews of trust in AI and human--automation
trust further suggest that trust depends on both system properties
and the user's interpretation of agency, reliability, and risk
~\cite{madhavan2007trust,glikson2020trust}. However, when agents
are embedded in social relationships, the issue is not limited to
whether users rely on them. Agents may also affect interpersonal
boundaries, social norms, and the distribution of responsibility.

In socially embedded HAI, trust, blame, and responsibility may be
distributed across users, agents, developers, institutions,
bystanders, and broader social contexts. This makes it necessary to
consider not only whether an agent is trusted, but also how the
agent is treated as a social participant when it refuses, avoids,
asserts a boundary, or becomes involved in socially disruptive
interaction.

\section{Acceptance and Avoidance as Relational Signals}



Human acceptance and avoidance do not emerge as instantaneous
binary reactions. They are grounded in temporally developing
cognitive, affective, and bodily states. In human interaction,
discomfort toward another person's approach, hesitation before
rejection, or the decision to continue a relationship often involves
gradual changes in attention, arousal, memory, expectation, and
interpretation. Such changes take time. Therefore, if a social agent
expresses acceptance, avoidance, discomfort, or rejection as an
immediate symbolic output, the expression may appear behaviorally
convenient but psychologically unnatural.

This view is consistent with process-oriented theories of emotion.
Scherer's component process model treats emotion as a dynamic
process of sequential appraisal checks, rather than as a single
instantaneous output~\cite{scherer2001appraisal}. Gross's process
model of emotion regulation also distinguishes multiple points at
which emotion can be modulated, including attention, cognition,
and response expression~\cite{gross1998emotionregulation}.

This point is important for human--agent interaction. An agent that
is always accepting may appear polite, but it lacks a believable
boundary. Conversely, an agent that immediately rejects or displays
anger may break the interaction before a relationship can be
adjusted. A more natural design requires an internal process in
which negative affect accumulates, is temporarily suppressed, leaks
through external behavior, and is eventually expressed depending
on the situation. The external expression should not be identical to
the internal state. Rather, the gap between the internal state and
the displayed behavior is itself a social signal. It allows an agent to
have something like ``honne'' and ``tatemae'': an internal state that
may differ from its outward expression, while still allowing the agent
to express rejection or avoidance according to the relationship and
the situation.

This paper extends our recent discussion of negative aversion and
real reliability, where discomfort and rejective reactions were framed
as legible self-disclosure for boundary learning and relational
repair~\cite{yonezawa2026negativeaversion}. Here, we shift the focus
from familiar reliability to trust and blame in socially risky
intermediation.

\section{From Avoidance to Affective Viscosity}

In this paper, \textit{affective viscosity} refers to the tendency
of negative affect to accumulate and persist over time,
characterized by its accumulation, retention, and decay.
These dynamics are influenced by the suppression and expression
of negative affect, and are particularly relevant to persistent
negative emotions such as jealousy.

Affective viscosity is therefore not simply a delay before responding.
An agent may retain discomfort while displaying restraint, respond
differently to the same event after repeated encounters, or express
an accumulated state abruptly after a tolerance threshold is exceeded.
The distinction between internal state and external expression is
central: changes in one need not be immediately mirrored in the other.

Here, affective viscosity is a design concept organizing these
processes, rather than a single empirically validated measure.
Acceptance and avoidance are also social effects. They do not only
represent whether the agent likes or dislikes the current user input.
They regulate distance, timing, intimacy, and the distribution of
attention among participants. 
When an agent accepts an approach, withdraws from it, tolerates it, or expresses discomfort toward it, the agent is not merely displaying an internal state; it is shaping the relationship.

This relational view is grounded in classic work on interpersonal
distance and relationship development. Hall's proxemics treats
distance as a socially meaningful organization of interpersonal
space~\cite{hall1966hidden}, while social penetration theory
describes relationship development as a gradual process of increasing
interpersonal depth~\cite{altman1973social}. Expectancy violations
theory further suggests that deviations from expected interpersonal
behavior can be evaluated positively or negatively depending on the
relationship and context~\cite{burgoon1993expectations}.

This view extends the issue from dyadic approach interaction to
broader social relationships. In approach-based interaction, the
design question is whether an agent should accept the user's
approach, endure it for a while, reject it, or remain involved despite
annoyance. Such behavior can support relationship maintenance, but
it can also become intrusive or inappropriate. In multi-party
interaction, acceptance and avoidance further involve how the agent
understands and expresses the relationship it wants to maintain.
For example, an agent may display jealousy or possessiveness not
as a simple negative emotion, but as an expression of the social
relationship it tries to secure.

Thus, affective viscosity should be understood not only as the
temporal persistence of negative affect, but also as the persistence
of social involvement. The agent's negative state may remain because
the relationship remains unresolved. Its avoidance may express a
boundary, while its persistence may express attachment. This
perspective prepares the ground for discussing social agents that
intermediate in human relationships, including the difficult question
of whether such agents can accept blame when their involvement
changes the social situation.

\section{Design Cases from Our Recent Studies}

\begin{table}[t]\scriptsize
\caption{Design cases connecting acceptance, avoidance, and affective viscosity.}
\label{tab:cases}
\begin{tabular}{p{0.1\linewidth}p{0.18\linewidth}p{0.25\linewidth}p{0.3\linewidth}}
\toprule
Case & Boundary or relation & Process or evaluation focus 
& Design implication \\
\midrule
Fujino et al. \cite{fujino2026haipos}& Personal-space intrusion & Stress accumulation and ``patience'' before anger & Rejection can be staged through suppressed expression. \\
Sunasaka et al. \cite{sunasaka2026haipos}& Inappropriate utterances & Discomfort accumulation and delayed escalation & Boundary expression needs temporal justification. \\
Hachiken et al. \cite{hachi2026haipos}& Approach distance and stay duration & Acceptance of proximity and continued presence & Intrusiveness depends on distance, duration, and relationship. \\
Obata et al. \cite{obata2026haipos}& Triadic interpersonal distance & Accumulated jealousy and avoidance & Negative affect can express relational investment. \\
\bottomrule
\end{tabular}
\end{table}

Building on our recent synthesis of negative aversion and real
reliability~\cite{yonezawa2026negativeaversion}, this section uses
four recent HAI poster studies as concrete design cases. Table~\ref{tab:cases}
summarizes how these cases connect acceptance, avoidance, discomfort,
stickiness, and jealousy to internal processes and design implications.
The following studies provide separate evaluations of particular
behaviors and user impressions. They are used here as empirical
design cases, rather than as a combined validation of affective
viscosity or of an agent's capacity to bear responsibility.

\subsection{Acceptance and Avoidance in Boundary-Sensitive Interaction}

This subsection follows our earlier prototype of approach-triggered
avoidance behavior, in which a robot accumulated and decayed
``dislike'' according to a user's physical approach, displayed
endurance below a tolerance threshold, and performed avoidance
when that threshold was exceeded~\cite{yonezawa2025avoidance}.

Our three recent studies examine the expression and perception
of acceptance, avoidance, and discomfort in boundary-sensitive
human--agent interaction.

\paragraph{Patience and suppression}
Fujino
et al. \cite{fujino2026haipos} proposed an emotion-suppressing agent that responds to
personal-space intrusion. In their model, the agent's internal stress
increases as the user approaches and decreases as the user moves
away, and the agent transitions among neutral, ``patience'', and
anger states. Here, ``patience'' is not a simple positive expression,
but a temporary suppression of unpleasant emotion before anger.
The evaluation comprised two online studies with 24 participants
each, using 0--99 visual analogue scales.
The first used static images to select a facial expression perceived
as ``patience''. The second used prerecorded approach videos to
evaluate impressions of the agent, including emotion suppression,
likability, and willingness to continue interacting.

Their evaluation showed that displaying ``patience'' before anger
increased perceived emotion suppression, willingness to continue
interacting, and likability. This case suggests that rejection or
avoidance can be staged through a gap between the internal state
and the external expression.

\paragraph{Patience in utterance}
Sunasaka et al. \cite{sunasaka2026haipos} extended this issue to voice interaction. Their
system estimates a Discomfort Index from inappropriate user
utterances and accumulates it with an emotional decay factor. The
agent then changes its vocal tone according to the accumulated
state. 
The pilot evaluation involved 24 participants who listened to
prerecorded, scripted audio dialogues while viewing text
transcripts. Ratings on 0--99 visual
analogue scales included approachability, trustworthiness, perceived
agent discomfort, emotionality, contemplation before speaking, and
willingness to interact. Thus, the evaluation examined responses to
controlled dialogue stimuli rather than live adaptation by the
proposed system.

Their pilot evaluation suggests that the policy of expression
is more important than the acoustic transition pattern alone: a
delayed escalation policy, in which the agent first gives a neutral
warning and expresses discomfort only after repeated inappropriate
utterances, was evaluated more positively than immediate boundary
setting. This case shows that avoidance and discomfort expression
should be designed as temporally justified responses rather than
as immediate negative outputs.

Because the escalation policies also differed in the number of
conversational turns, the results do not isolate the effect of delay
alone.

\paragraph{Acceptance of approach and continued presence}
Hachiken et al. \cite{hachi2026haipos} examined a related boundary problem through
spatial and temporal involvement. 
Although the study uses the term spatiotemporal stickiness,
its current experimental contribution is best understood as an
evaluation of how users accept or reject an agent's approach and
continued presence.

Their within-subject robot interaction study involved 24 participants
and manipulated approach distance (30 or 50 cm), stay duration
(20 or 40 s), and relational framing (stranger, acquaintance, or
close friend). Seven-point ratings
assessed friendliness, favorability, discomfort, annoyance,
enjoyment, and continued interaction intent. Relationships were
introduced through instructions for each trial rather than developed
through long-term interaction.

The results showed that excessive proximity
increased discomfort and annoyance, whereas in close relationships,
brief and distant contact could also be evaluated negatively. This
case suggests that acceptance is not determined only by whether
the agent approaches the user, but also by how long it remains,
how close it comes, and what relationship is assumed.

\subsection{Affective Viscosity in Social Relationships}

Obata et al. \cite{obata2026haipos} provide a case in which negative affect is generated
from a social relationship rather than from a dyadic boundary
violation alone. 
Their VR agent extends Heider's POX model of balance theory into
a distance-based continuous model among the agent, a third party,
and the object of affection~\cite{heider1958psychology}.
The agent independently accumulates
two discomfort values: jealousy, caused by the third party's presence,
and avoidance, caused by being too close to the object. When the
accumulated value exceeds a threshold, the agent performs a
resolution behavior to restore the desired distance.

This case is important because negative affect is not treated as a
momentary reaction to a user's direct behavior. Instead, it emerges
from the discrepancy between the agent's ideal and actual social
configuration. 

The within-subject VR study involved 25 participants and compared
a two-party model with two three-party models differing in the
agent's desired relative distance~\cite{obata2026haipos}.
Seven-point questionnaire ratings assessed competitiveness,
familiarity, attachment, human-likeness, and jealousy-related
emotions, supplemented by free descriptions.

The evaluation showed that the three-party model,
especially the condition in which the agent wanted to be closer to
the object than the third party, increased impressions related to
competitiveness, possessiveness, familiarity, and attachment. Thus,
jealousy can be understood not only as a negative emotional display,
but also as an expression of the relationship that the agent attempts
to secure.

In this sense, the jealousy model provides a core example of
affective viscosity in social relationships. The negative state is
generated by relational imbalance, accumulates over time, and is
released through behavior that changes the social configuration.
This extends the design problem from whether an agent should
accept or avoid a user's approach to how an agent maintains,
protects, or modifies a relationship involving multiple social actors.

\begin{figure*}[t]
  \centering
\includegraphics[width=.45\linewidth]{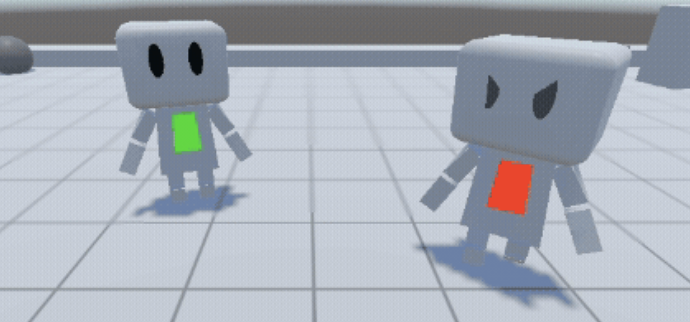}
  \hfill
\includegraphics[width=.45\linewidth]{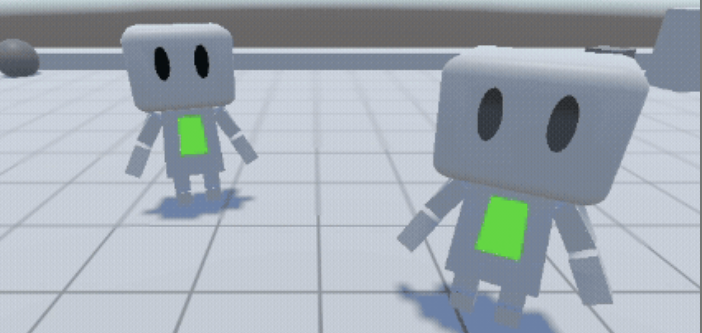}
 \begin{small}
 \begin{minipage}{0.45\linewidth}
A: The agent on the right grew jealous and angry when the agent on the left approached the user.
 \end{minipage}  \hfill
  \begin{minipage}{0.45\linewidth}
B: The right agent then moved much closer to the user, gaining satisfaction from the difference in distance. 
 \end{minipage} 
 \end{small}
 \caption{Triadic jealousy agent based on interpersonal distances involving the user.}
 \label{fig:obataPoster}
\end{figure*}

Figure~\ref{fig:obataPoster} illustrates the triadic situation used by
Obata et al. The agent accumulates jealousy when a third party
approaches the user and then performs a resolution behavior by
moving closer to the user. This figure is used here as a concrete
example of how negative affect can be generated from a social
configuration rather than from a dyadic boundary violation alone.

\section{Discussion}


\begin{figure}[t]
  \centering
\begin{picture}(\linewidth,160)(-0.3\linewidth,-150)
  \put(-70, 0){\makebox(0,0)[l]{\fbox{\parbox{8cm}{\centering Acceptance / Avoidance}}}}
  \put(45, -15){\makebox(0,0)[l]{$\downarrow$}}
  \put(-70, -35){\makebox(0,0)[l]{\fbox{\parbox{8cm}{\centering Boundary-sensitive interaction \\ (personal space, utterance, approach)}}}}
  \put(45, -55){\makebox(0,0)[l]{$\downarrow$}}
  
  \put(-70, -70){\makebox(0,0)[l]{\fbox{\parbox{8cm}{\centering Internal accumulation / suppression / delayed expression}}}}
  
  \put(45, -85){\makebox(0,0)[l]{$\downarrow$}}
  \put(-70, -100){\makebox(0,0)[l]{\fbox{\parbox{8cm}{\centering Affective viscosity}}}}
  \put(45, -115){\makebox(0,0)[l]{$\downarrow$}}
  \put(-70, -135){\makebox(0,0)[l]{\fbox{\parbox{8cm}{\centering Social relationship regulation\\(triadic jealousy, persistence, blame)}}}}
  %
\end{picture}
\caption{Conceptual Models of Agents' Acceptance, Avoidance, and Socially Affective Viscosity}\label{fig:concept}
\end{figure}

The cases discussed above suggest that social agents do not merely
respond to users; they intermediate in relationships. 
When an agent expresses patience, discomfort, or jealousy,
or behaves in ways that users find annoying, it may clarify a
boundary, preserve a relationship, or reveal a social tension
that had remained implicit.
Such behavior may improve the social environment, but it may also make the agent disliked or blamed.

The issue of blame is important here because blame is not a simple
emotional reaction after failure, but a structured judgment involving
causality, agency, intentionality, and obligation~\cite{malle2014blame}.
In socially embedded HAI, such judgments may become distributed
or fragmented across users, agents, developers, institutions, and
surrounding social contexts. The question is therefore not only who
is blamed after a failure, but how an agent can be designed to take
part in a socially risky situation before the problem worsens.

Figure~\ref{fig:concept} summarizes this argument. Acceptance and
avoidance first appear in boundary-sensitive interaction, such as
personal space, utterances, and approach. These responses then
require internal accumulation, suppression, and delayed expression,
which we describe as affective viscosity. Finally, affective viscosity
leads to social relationship regulation, including triadic jealousy,
persistence, and blame.

From this perspective, making an agent responsible does not mean placing an artificial scapegoat or villain in the environment. Rather, it means designing the agent's involvement so that it is traceable, limited, contestable, and repairable. Traceability means that users can understand why the agent acted. Limitation means that the agent's authority and scope of involvement are bounded. Contestability means that users can reject, question, or override the agent's behavior. Repairability means that the agent can withdraw, explain, apologize, or change its behavior after a failed involvement.

Hachiken et al.'s notion of spatiotemporal stickiness points toward
a longer-term issue: an agent's involvement may become acceptable,
annoying, or blameworthy not only because it approaches too closely,
but because it remains, returns, or repeatedly tries to maintain the
relationship~\cite{hachi2026haipos}. Similarly, Obata et al.'s
jealousy model suggests that negative affect may express relational
investment, while also creating a risk that the agent disturbs or
exposes the relationship~\cite{obata2026haipos}. These cases
suggest that blame should not be treated only as a post-hoc
attribution after failure. It should also be considered as a design
problem for agents that intermediate in socially risky relationships.
Although ``honne'' and ``tatemae'' offer one way to describe
the gap between internal states and outward expression,
the interpretation of such expressions is culturally situated.
Suppressed discomfort or delayed rejection may be perceived
as considerate restraint or as passive aggression, while
jealousy may suggest attachment or unwanted intrusion.
The social consequences of emotion suppression can differ
across cultural contexts~\cite{butler2007culture}.
Our cases do not test these differences. Future work should
examine how interpersonal distance, communication norms,
and relational expectations shape acceptance and blame attribution.

\section{Conclusion}
This paper connected four recent studies on personal-space
intrusion, vocal discomfort, spatiotemporal stickiness, and triadic
jealousy to the broader question of how social agents intermediate
in human relationships. We argued that acceptance and avoidance
should not be treated as immediate binary outputs, but as temporally
developing internal states and social effects. Through the notion of
affective viscosity, negative states such as discomfort, reluctance,
annoyance, and jealousy can be understood as states that accumulate,
are suppressed, leak through behavior, and become expressed
according to relationship and situation.

The main question raised by this paper is how agents can bear
responsibility for such involvement. When agents express boundaries
or become involved in social tension, they may support relationship
regulation, but they may also become targets of dislike or blame.
Future work should therefore examine how agents can intermediate
in socially risky relationships while making their involvement
traceable, limited, contestable, and repairable.

\begin{acks}
This research was supported in part by JSPS KAKENHI
(Grant Numbers 23K11278, 26K14911, 23K11202, and 24K02977)
and the JST Moonshot R\&D Program (Grant Number JPMJMS2215).
JPMJMS2215 supported the discussion of the ethical and social
implications of negative affect expression by social agents.
The author would like to thank the students in our laboratory who contributed to the four HAI 2026 studies discussed in this paper: Atsuo Fujino, Tatsuyoshi Sunasaka, Hiromu Hachiken, and Kuma Obata as the first authors, and Tomoki Yanagi and Takaya Tamamoto as co-researchers.
\end{acks}

\end{document}